%% file: aaai21-biobjective-classification-trees.tex
\def\year{2021}\relax
\documentclass[letterpaper]{article} 
\usepackage{aaai21}  
\usepackage{times}  
\usepackage{helvet} 
\usepackage{courier}  
\usepackage[hyphens]{url}  
\usepackage{graphicx} 
\usepackage{graphicx}  
\usepackage{natbib}  
\usepackage{caption} 
\usepackage{color} 
\usepackage{amsmath, amsthm, amssymb}
\usepackage[switch]{lineno}
\usepackage{mathtools}

\newtheorem{definition}{Definition}

\newcommand{\ignore}[1]{}

\newcommand{\nondom}{\mathit{nondom}}
\newcommand{\dom}{\triangleleft}

\newcommand{\citeay}[1]{\citeauthor{#1}~(\citeyear{#1})}

\title{Optimal Decision Trees for Nonlinear Metrics}

\title{Optimal Decision Trees for Nonlinear Metrics}

\author {
        Emir Demirovi\'{c},\textsuperscript{\rm 1}
        Peter J. Stuckey \textsuperscript{\rm 2}\\
}
\affiliations {
    \textsuperscript{\rm 1} Delft University of Technology, The Netherlands \\
    \textsuperscript{\rm 2} Monash University and Data61, Australia \\
    e.demirovic@tudelft.nl, peter.stuckey@monash.edu
}

\begin{document}
\maketitle

\begin{abstract}\footnote{This article has been published in AAAI'21: \url{https://ojs.aaai.org/index.php/AAAI/article/view/16490}}
Nonlinear metrics, such as the F1-score, Matthews correlation coefficient, and Fowlkes–Mallows index, are often used to evaluate the performance of machine learning models, in particular, when facing imbalanced datasets that contain more samples of one class than the other. Recent optimal decision tree algorithms have shown remarkable progress in producing trees that are optimal with respect to linear criteria, such as accuracy, but unfortunately nonlinear metrics remain a challenge. To address this gap, we propose a novel algorithm based on bi-objective optimisation, which treats misclassifications of each binary class as a separate objective. We show that, for a large class of metrics, the optimal tree lies on the Pareto frontier. Consequently, we obtain the optimal tree by using our method to generate the set of all nondominated trees. To the best of our knowledge, this is the first method to compute provably optimal decision trees for nonlinear metrics. Our approach leads to a trade-off when compared to optimising linear metrics: the resulting trees may be more desirable according to the given nonlinear metric at the expense of higher runtimes. Nevertheless, the experiments illustrate that runtimes are reasonable for majority of the tested datasets.
\end{abstract}

\section{Introduction}

Decision trees are amongst the most explainable machine learning models.
There has been recent interest in building trees that, given a dataset, are optimal with respect to some metric,
e.g., trees that minimise misclassifications \cite{dl8} or sparse trees that trades off misclassifications for
size~\cite{bertsimas2017optimal}. In the following, we focus on binary classification using decision trees.

Metrics considered in previous works are linear, 
i.e., the objective is a linear sum over the misclassifications of each class.
This allows solving the problem by separately optimising the left and right subtree and joining the results.
Such a decomposable structure is key in efficient algorithm design and is one of the main
advantages of dedicated decision tree approaches over general-purpose optimisation methods.

Nonlinear metrics, such as the F1-score, establish a nonlinear relationship between the misclassifications of each class.
This is preferred over linear metrics in many applications, especially if the dataset is heavily skewed towards one class.
For example, a trivial classifier that classifies all individuals as \emph{healthy} may result in high accuracy since most
data points correspond to healthy individuals, but bears no significance in detecting early signs of an illness.
In these cases, a nonlinear metric may be more appropriate.

The challenge with nonlinear metrics is that foundational principles of previous works, which were important for the success of the methods, no longer hold. Dedicated approaches rely on the independence of the left and right subtree, which is not the case for nonlinear metrics.
Integer programming cannot be employed as the objective is nonlinear. 
Therefore, new techniques are required.

We propose a novel approach to address the problem of nonlinear metrics by viewing the problem as a bi-objective problem,
where the misclassifications of each binary class corresponds to one objective. 
We show that, for most machine learning metrics, the optimal classifiers lie on the Pareto frontier,
i.e., the set of nondominated trade-offs between the two objectives.
After computing all Pareto optimal trees, the optimal tree may be selected according to the desired metric.
Our bi-objective algorithm is able to take advantage of the unique structure of decision trees by generalising
previous dedicated decision tree approaches to the bi-optimisation setting, using techniques based on dynamic programming and search.
The bi-objective view allows us to optimise for a wide range of metrics at the expense of runtime and provides a trade-off:
linear metrics are easier to optimise but may be less appropriate for a given application, whereas our approach provides the provably optimal tree for any desired metric albeit a higher runtime.

The rest of the paper is organised as follows. In the next section, we discuss preliminary notions and give details on previous decision tree algorithms. Afterwards, we describe our approach, which consists of showing that Pareto optimal trees lie on the frontier, generalising decision tree concepts to the bi-objective case, and generalising algorithmic enhancement that improve the efficiency of the algorithm. The value of our method is given in a dedicated experimental section, where we consider 75 publicly available datasets. We end by discussing related work and give a conclusion.

\section{Preliminaries}
\label{section:preliminaries}

An \emph{instance} is a pair $\mathbf{I} = (\mathbf{fv}, class)$, where $\mathbf{fv} \in \mathbb{R}$ is a feature vector and $class \in \mathbb{N}$. We consider binary classification problems, i.e., $class \in \{0, 1\}$. We denote the feature vector and class of an instance $\mathbf{I}$ as $\mathbf{fv}(\mathbf{I})$ and $class(\mathbf{I})$, respectively. A \emph{dataset} is a set of instances. Datasets $\mathcal{D}^+$ and $\mathcal{D}^-$ are the sets of positive and negative instances. A \emph{binary classifier} is a mapping $\mathbf{f}: \mathbf{fv} \rightarrow \{0, 1\}$. Given a dataset $\mathcal{D}$ and a binary classifier $\mathbf{f}$, an instance $\mathbf{I} \in \mathcal{D}$ is a \emph{false negative} if the classifier $f$ incorrectly classifies instance $\mathbf{I}$ as a negative class, i.e., $f(\mathbf{fv}(\mathbf{I})) = 0$ and $class(\mathbf{I}) = 1$. \emph{False positives} are defined analogously. \emph{Learning} a classifier corresponds to selecting a function $\mathbf{f}$ from a family of functions aiming to maximise a given metric. We consider learning decision trees (see further).


Decision trees are binary trees, where each node is either a \emph{predicate
  node}, which is assigned a predicate, or a \emph{classification
  node} (\emph{leaf node}), which is assigned a class. A classification tree classifies an
instance according to the following recursive procedure starting with the
root node: if the node is a classification node, return the class assigned
to the node, otherwise recurse on the left or right child node depending on
the result of applying the instance to the node predicate. A common
predicate choice is to evaluate whether a
particular feature of the instances exceeds a threshold. The \emph{depth} of
a tree is maximum number of predicates from the root node to any leaf
node. We define the \emph{size} of a tree as the number of predicate nodes in the tree. Note that a tree with $n$ predicate nodes has $n+1$ leaf nodes.

In our setting, the set of possible predicates are given upfront. As a result, instances are represented as \emph{binary} feature vectors, where each feature corresponds to the outcome of a predicate. We say a feature $f_i$ or its negation $\overline{f_i}$ is in the feature vector if the \emph{i-th} feature is true or false. The set of features is denoted as $\mathcal{F}$. As predicates operating on binary features are identify functions, we say that a (binary) feature, rather than a predicate, is assigned to a predicate/feature node. For a dataset $\mathcal{D}$, we denote $\mathcal{D}(f)$ the set of instances that contain feature $f$.

In light of the previous discussion, given a dataset which contains nonbinary features, the predicates must be determined through a \emph{binarisation} process of the dataset. Note that every decision tree algorithm, implicitly or explicitly, binarises feature vectors through predicates. In future text, we assume binary datasets are given.

We consider the following metrics for evaluating decision trees. Linear
metrics: 
\emph{Accuracy} is the ratio between correctly classified instances 
and the size of the dataset, equivalently the \emph{misclassification score}
is the number of misclassifications. \emph{Weighted accuracy/misclassifications} weighs individual classes differently, e.g. $w_p \times fp + w_n \times fn$ for
weight $w$. 
Nonlinear metrics establish a nonlinear relationship between true positives
(\emph{tp}), true negatives (\emph{tn}), false positives (\emph{fp}), and
false negatives (\emph{fn}): \emph{F1-score}: $tp / (tp + 0.5(fp+fn))$,
\emph{Matthiews correlation coefficient}: $(tp \times tn - fp \times fn) /
\sqrt{(tp+fp)(tp+fn)(tn+fp)(fn+fn)}$, \emph{Fowlkes-Mallows index}:
$\sqrt{tp^2 / ((tp+fp)(tp+fn))}$. For datasets with one dominant class,
e.g., detecting illness amongst a mostly healthy population, nonlinear
metrics may be more appropriate, since even trivial classifiers achieve
seemingly high accuracy but do not carry useful information.
To avoid overfitting, standard practice in machine learning is to partition the given dataset into a \emph{training} and \emph{testing} set, and then construct a tree using the training set, and evaluate the quality on the testing set.

Bi-objective optimisation aims to simultaneously optimise two (competing)
objective functions with respect to a set of constraints. Solutions are
represented as pairs $(x, y)$, where $x$ and $y$ are the objective values
of the two objectives, respectively.
\ignore{For the minimisation case (maximisation is analogous), objective $o_1 = (x_1, y_1)$ \emph{dominates} objective $o_2 = (x_2, y_2)$ iff $x_1 \leq x_2 \land y_1 \leq y_2$. \emph{Strict domination} requires $x_1 < x_2 \land y_1 \leq y_2 \lor x_1 \leq x_2 \land y_1 < y_2$. We say $(x_1, y_1)$ and $(x_2, y_2)$ are \emph{nondominating} if neither solution dominates the other. The \emph{Pareto front} is the set of all nondominated solutions.
}
Given a bi-objective problem $P$ where the aim is to minimize both
objectives 
we say that one solution $(a,b)$ \emph{dominates another} $(a',b')$, written
$(a,b) \dom (a'b')$, iff $a \leq a' \wedge b \leq b' \wedge (a,b)
\neq (a',b')$.
Given a set of pairs of objective values $S$ define:
$$
\nondom(S) = \{  (a,b) \in S ~|~ \neg\exists (a',b') \in S,  (a',b') \dom
(a,b) \}.
$$
The \emph{Pareto front} of a bi-objective problem $P$ with solutions $S$ is
exactly the set $\nondom(S)$. Given a set of pairs $U$ and another set of pairs $L$ we say that
$L > U$ if $\forall (a,b) \in L, \exists (a',b') \in U \text{~where~} (a',b') \dom
(a,b)$.

\subsection{Dynamic Programming and Search for Trees}

We describe previous work on dynamic programming and search for constructing
decision trees with minimal misclassifications
in more detail \cite{dl85,dl8,murtree}, as we generalise these methods.



A crucial property of a decision tree is its decomposable structure: the
misclassifications of a parent node is obtained as the sum of the
misclassifications of its children, allowing the left and right subtrees to
be optimised independently. 
This leads to the dynamic programming formulation to compute the minimum misclassification score $T(\mathcal{D}, d)$ for a dataset $\mathcal{D}$ and a tree of depth $d$ given the set of features $\mathcal{F}$:

\begin{equation}
\label{DPformulation}
T(\mathcal{D}, d) = 
\left\{ \begin{array}{ll}
\min\{|\mathcal{D}^+|, |\mathcal{D}^-|\} & d =0\\
 \min\{T(\mathcal{D}(\overline{f}), d-1)   & d > 0 \\
~~~~~+ T(\mathcal{D}(f), d-1) : f \in \mathcal{F}\} & 
\end{array}
\right.
\end{equation}

The base case in Eq.~\ref{DPformulation} defines the misclassification score for classification nodes. The general case states that computing the optimal misclassification score amounts to examining all possible feature splits. For each feature, the optimal misclassification is computed recursively as the sum of the optimal misclassifications of its children. See \cite{murtree} for a formulation that additionally allows constraining the number of nodes, however such a formulation is not necessary for understanding our main contributions. We proceed with techniques that speed-up the computation. 

\emph{Specialised bi-objective algorithm for trees of depth two
  \cite{murtree}}. Optimal decision trees of depth two may be constructed
based on the pairwise-feature frequency counts, leading to a more efficient
method for depth two trees compared to a direct applications of
Eq. \ref{DPformulation}. The positive/negative frequency count $FQ^{+/-}$ of
two (possibly negated) features is given as the number of positive/negative
instances that contains both features. Given a tree of depth two, a root
node with feature $f_{root}$, and a right child node with feature
$f_{right}$, the misclassification score of the right-most classification
node can be computed as $\min\{FQ^+(f_{root}, f_{right}), FQ^-(f_{root},
f_{right})\}$. The misclassification score of the other classification nodes
can be computed analogously. Iterating through all pairs of features allows
us to compute the best left and right subtrees for each feature, and then
the optimal tree is selected as the tree that minimises the
misclassification score. 
The benefit of this approach is that it requires work proportional to
square of the maximum number of features occuring in an instance, rather
than the square of the total number of feature (see~\cite{murtree} for details).

\emph{Caching \cite{dl8}}. A subtree may be considered multiple times when using Eq. \ref{DPformulation}. For example, consider the depth three right-most subtree with $f_1$ and $f_2$ as the features of the root and right child nodes. Swapping $f_1$ and $f_2$ does not change the considered subtree. Following this observation, each optimal subtree is stored in a cache, and when a new subtree is considered, computation may be avoided if the subtree has already been stored in the cache.

\emph{Pruning based on lower and upper bounds}. For each node an upper and lower bound is computed. If the lower bound exceeds the upper bound, the node is pruned since further processing it cannot lead to an improving tree. Initially trivial bounds are used for nodes and are refined during the search as follows. Bounds by \cite{dl85}: (A) Given a parent node, its optimal left subtree with misclassification cost $\alpha$, and the best incumbent for the parent node found so far is $\beta$ misclassifications, then an upper bound of $\beta - \alpha - 1$ misclassifications may be imposed on the right subtree, and (B) if no tree with $UB$ misclassifications could be found, then $UB+1$ is stored in the cache as the lower bound for the subtree. Bounds by \cite{murtree}: (C) a lower bound for a node can be computed by considering every possible split and computing the sum of the lower bounds of the left and right subtrees and taking the minimum, and (D) similarity-based lower bounding computes a bound for a given dataset $\mathcal{D}_2$ based on the optimal misclassification score $\alpha$ of another dataset $\mathcal{D}_1$ using set difference operations. The intuition is as follows: optimistically assume that instances $\mathcal{D}_2 \setminus \mathcal{D}_1$ will be perfectly classified and pessimistically assume that each instance from $\mathcal{D}_1 \setminus \mathcal{D}_2$ was misclassfied. Based on this extreme reasoning, we arrive at a misclassification bound $\alpha - |\mathcal{D}_1 \setminus \mathcal{D}_2|$ for $\mathcal{D}_2$.

\ignore{
A summary of the algorithm is given in Algorithm X. [maybe we can combine
  the pseudo-code of the single- and bi-objective versions]
}
\section{Algorithm for Pareto Optimal Decision Trees}

Nonlinear metrics, such as the F1-score, are a challenge for decision tree algorithms since these types of metrics cannot be optimised using the dynamic programming formulation (Eq. \ref{DPformulation}). Notably, it no longer holds that the parent node can be solved by independently optimising its child nodes and summing the results. To address nonlinear metrics, we propose an approach based on bi-objective optimisation. This leads to a generic approach that benefits from exploiting the decision tree structure.

We consider classification metrics of the form $f(fp, fn)$, where \emph{fp} and \emph{fn} are the number of false positives and false negatives, that satisfy the following montonicity property:

\begin{gather}
\label{eq:monotonicity}
\forall fp, fn, fp', fn' \\ \nonumber fp \leq fp' \land fn \leq fn ' \implies f(fp, fn) \succ f(fp', fn')
\end{gather}
\noindent
where $\succ$ is $\geq$ (maximisation) or $\leq$ (minimisation). Intuitively, monotonicity states that given a classifier, it is always beneficial to further improves its classification. Most popular classification metrics satisfy Eq. \ref{eq:monotonicity}, e.g., accuracy, F1-score, Matthiews correlation coefficient. We are not aware of practical metrics that do not satisfy Eq. \ref{eq:monotonicity}.

The key observation is that Eq. \ref{eq:monotonicity} implies the optimal
point for $f(fp, fn)$ is a point on the Pareto front, the set of
nondominated trade-offs between the number of false positives and false
negatives (proof by contradiction). Therefore, optimising $f(fp, fn)$ may be
done by computing the Pareto front and selecting the best tree according to
the metric. Note that while such an approach works for any monotonic metric, linear metrics such as misclassifications 
are best optimised with a less expensive single-objective approach.

To the best of our knowledge, there is no other approach that can produce provably optimal trees with respect to nonlinear metrics. \cite{icmlTree} noted the difficulty of the F1-score and proposed to linearise the problem by changing the node classification criteria, however this does not address the optimisation problem. In remaining text, we describe our bi-objective tree algorithm for computing the Pareto front.

\subsection{Bi-Objective Optimisation}

We start by lifting definitions from the single- to the bi-objective case to arrive at the bi-objective dynamic programming formulation to compute the Pareto front, and afterwards discuss additional algorithmic techniques. 

\begin{definition}(Bi-Misclassification)
The bi-misclassification of a tree $T$ on a dataset $\mathcal{D}$ is a pair $(fp, fn)$, where $fp$ and $fn$ are the number of false positives and false negatives misclassified by the tree $T$ on the dataset $\mathcal{D}$. 
\end{definition}

The decomposable structure of decision trees remains important in the bi-objective case, but when combining the optimal result of the left and right child, a generalised merge operation needs to be employed: 


\begin{definition}(Merge)
\label{definition:merge}
Given a parent node $p$ with feature $f$ and Pareto fronts of its children
$PF_{left}$ and $PF_{right}$, the Pareto front $PF_{p, f} =
merge(PF_{left},PF_{right})$ 
of the parent node $p$ with feature $f$ is computed as $PF_{p, f} = \nondom(\{ (x_1 + x_2, y_1 + y_2) : (x_1, y_1) \in PF_{left} \land (x_2, x_2) \in PF_{right}\})$.
\end{definition}

We now state our bi-objective dynamic programming formulation, which provides a high-level view of our algorithm.  Given a dataset $\mathcal{D}$ and the depth $d$ of the tree, the Pareto front can be computed as follows:

\begin{equation}
\label{BiDPformulation}
T(\mathcal{D}, d) = 
\left\{ \begin{array}{ll}
\{(|\mathcal{D}^+|, 0), (0, |\mathcal{D}^-|)\} & d =0\\
 \nondom(\bigcup_{ f \in \mathcal{F}} merge(   & d > 0 \\
T(\mathcal{D}(\overline{f}), d-1), T(\mathcal{D}(f), d-1))]) &
\end{array}
\right.
\end{equation}

Eq. \ref{BiDPformulation} generalises Eq. \ref{DPformulation} to the bi-objective case. The base case, representing a leaf node ($d=0$), returns a Pareto front representing two classification nodes, one for each class. The general case is similar to the single-objective version (Eq. \ref{DPformulation}), but the general merge operation (Def. \ref{definition:merge}) is performed to aggregate the Pareto fronts of the child nodes, and the filter operator is used to remove dominated points. Limiting the size of the tree is possible as in the single-objective case (see \cite{murtree} for details), but we do not discuss it for brevity of presentation.

\subsection{Algorithmic Enhancements}

We generalised several algorithmic components to the bi-objective case to improve the computational efficiency when implementing Eq. \ref{BiDPformulation}, namely the specialised bi-objective algorithm for trees of depth two, similarity-based lower bounding, and caching overlapping subproblems.  Note that caching optimal subtrees is the same as in the single-objective case.

\emph{Upper and Lower bounds}. Each node is assigned an upper bound and lower bound, which are in our bi-objective setting a set of nondominating pairs of integers rather than a single value as in the single-objective case. Initially, the upper and lower bound of the root node is set to the trivial bounds $\{(\infty, \infty) \}$ and $\{(0, 0)\}$, respectively, and will be refined during the algorithm. The bounds play a role in defining infeasibility and pruning.

\begin{definition}(Bi-Objective Lower Bound)
\label{def:lb}
Given a dataset $\mathcal{D}$, a set of nondominated pairs of integers is a
lower bound $LB$ 
if for every decision tree there is at least one element of the lower bound
that dominates or equals it.
\end{definition}

\begin{definition}(Bi-Objective Upper Bound)
\label{def:ub}
Given a dataset $\mathcal{D}$, a set of nondominated pairs of integers is an
upper  bound $UB$ if for every decision tree 
there is at least one element of the upper bound it dominates or is equal to.
\end{definition}

\begin{definition}(Infeasible Node)
Given a dataset $\mathcal{D}$ and a node, a lower bound $LB$, and an upper
bound $UB$, the node is called $infeasible$ if every element of the upper
bound is dominated by at least one element of the lower bound, i.e., $LB > UB$.
\end{definition}

Infeasible nodes are pruned during the search since expanding them cannot produce any tree that belong to the optimal Pareto front.

We may now present the generalisation of the techniques to compute upper and lower bounds (see preliminaries).

(A) Upper bound computation. Given a parent node with $UB$ as an upper bound
on its solutions, 
and an optimal left subtree with Pareto front $PF_{left}$, 
an upper bound for the right subtree may be computed as: 

\begin{gather}
\label{eq:upperBounding}
UB_{right} = \nondom(\{(x_1 - x_2, y_1 - y_2) : \nonumber \\ (x_1, y_1) \in UB \land (x_2, y_2) \in PF_{left} \}).
\end{gather}

(B) Lower bounds based on infeasibility. After exhaustively exploring a node, if no tree that dominates an element of the upper bound $UB$ could be computed, then $UB$ is stored in a cache as a \emph{lower bound} for the subtree. Note that in case at least one solution was computed, the lower bound is not valid in general according to Def \ref{def:lb}. 

(C) Combining the lower bounds of child nodes. 
A lower bound for the depth $d> 0$ subtree on the dataset $\mathcal{D}$ may
be computed 
by \emph{lookahead} as follows:

\begin{equation}
\begin{multlined}
\label{LBformulation}
LB_{look}(\mathcal{D}) = \\ \nondom(\bigcup_{ f \in \mathcal{F}} merge(LB(\mathcal{D}(\overline{f})), LB(\mathcal{D}(f)))
\end{multlined}
\end{equation}
\noindent where $LB$ is any other lower bounding approach that is applicable.

(D) Similarity-based lower bounding. We extend the technique by reasoning on the maximum reduction for each objective. We may compute a lower bound $LB$ for a dataset $\mathcal{D}$ based on the Pareto front $PF_i$ of another dataset $\mathcal{D}_i$. Optimistically assume that each $\mathcal{D}^+ \setminus \mathcal{D}^+_i$ was misclassified by every tree on the Pareto front $PF_i$, i.e., $b^+ = |\mathcal{D}^+ \setminus \mathcal{D}^+_i|$ is the maximum reduction possible for the misclassifications of the positive class and analogously for $b^-$ and the negative class. The lower bound for $\mathcal{D}$ may be computed as applying the reduction on each element of the Pareto front $PF_i$:

\begin{equation}
\label{eq:lbsim}
LB_{sim}(\mathcal{D}) = \{(x - b^+, y - b^-) : (x, y) \in PF_i\}.
\end{equation}

\emph{Pareto front data structure and the Merge operator}. The Pareto front
is represented as an array sorted by the first objective. 
The Merge operator
(Def. \ref{definition:merge}) is implemented by lexicographically sorting
the the pairs $(a,b)$, and them removing dominated pairs in a linear pass:
the element $(a,b)$ removes all pairs until reaching a pair $(a',b')$ with
$b' < b$.

\emph{Specialised bi-objective algorithm for trees of depth two}. We
modified the specialised algorithm to use the Merge operator
(Def. \ref{definition:merge}) and use two elements of the Pareto front for
classification nodes in the base case as given in Eq. \ref{BiDPformulation}.
We denote the specialized algorithm as $T_2(\mathcal{D}, d)$ and extend it to
also handle depths 0 and 1 which are much easier.

\emph{Summary of the algorithm}. Pseudo-code for the algorithm is given in Figure~\ref{fig:code},
where details on bounding the size of the tree in terms of numbers of nodes 
are elided for simplicity. The call $T(\mathcal{D}, d, UB)$ returns a Pareto
front of (false positive, false negative) values possible for decision trees
for the dataset $\mathcal{D}$ with depth $d$ that are not dominated by
a pair in set $UB$.  The algorithm caches both solutions, as $\langle F,
optimal \rangle$,
and lower bounds, as $\langle F, lb \rangle$. An empty cache returns $\langle
\emptyset, \bot \rangle$.

\begin{figure}[t]
\begin{tabbing}
xx \= xx \= xx \= xx \= xx \= xx \= \kill
\> \> $UB_{right}$ := \= $nondom(\{ (x_1 - x_2, y_1- y_2)$ \= \kill
$T(\mathcal{D}, d, UB)$ \\
\> \textbf{if} ($d \leq 2$) \\
\> \> $F$ := $T_2(\mathcal{D},d)$ \\
\> \> $stat$ := $optimal$ \\
\> \textbf{else} $\langle{F, stat}\rangle$ := $cache[\mathcal{D},d]$ \\
\> \textbf{if} ($F > UB$) \textbf{return} $\emptyset$ \\ 
\> \textbf{if} ($stat = optimal$) \textbf{return} $F$ \\
\> $PF$ := $\emptyset$ \\
\> \textbf{for}($f \in \mathcal{F}$) \\
\> \> $LB_{left}$ := $LB(D(\overline{f}))$ \> \> \% (D)\\
\> \> $LB_{right}$ := $LB(D(f))$         \> \> \% (D) \\
\> \> $LB$ := $merge(LB_{left},LB_{right})$ \> \> \% (C) \\
\> \> \textbf{if} ($LB > UB$) \textbf{continue} \\
\> \> $PF_{left}$ := $T(D(\overline{f}), d-1, UB)$ \\
\> \> \textbf{if} ($PF_{left} = \emptyset$) \textbf{continue} \\
\> \> $UB_{right}$ := \= $nondom(\{ (x_1 - x_2, y_1- y_2)$ \> \% (A) \\
 \> \> \>  $|~ (x_1,y_1) \in UB, (x_2,y_2) \in PF_{left} \}$ \\
\> \> $PF_{right}$ := $T(D(f), d-1, UB_{right})$ \\
\> \> \textbf{if} ($PF_{right} = \emptyset$) \textbf{continue} \\
\> \> $F$ := $merge(PF_{left},PF_{right})$ \\
\> \> $PF$ := $nondom(PF \cup F)$ \\
\> \> $UB$ := $nondom(UB \cup F)$ \\ 
\> \textbf{if} ($PF > UB$) \\
\> \> $cache[\mathcal{D}, d]$ := $\langle PF, lb \rangle$ \> \> \% (B)\\
\> \> \textbf{return} $\emptyset$ \\
\> $cache[\mathcal{D},d]$ := $\langle PF, optimal \rangle$ \\
\> \textbf{return} $PF$    
\end{tabbing}
\caption{Pseudo-code for the bi-objective dynamic program
  $T$. Comments mark where bounds optimizations are applied.\label{fig:code}}
\end{figure}

The algorithm calls the specialized method for trees of depth less than 2,
otherwise it looks up the cache entry. 
If the resulting frontier $F$ (optimal or lower bound) is infeasible wrt to the upper
bound it returns $\emptyset$. 
If the result is optimal it returns it.
Otherwise it considers each feature $f$ in turn. It calculates the lower bound
for the tree split on this feature,
using the strongest available lower bounds on the left and right subtrees,
 if that is incompatible with the  
upper bound it moves to the next feature.
It then recursively solves the left child. If this has no solutions it 
skips to the next feature. Otherwise it refines the upper bound for the
right tree, and recursively solves the right child.  
Otherwise it merges the result and updates the pareto frontier $PF$
and the current upper bound $UB$. 
When all features are examined if the result is still incompatible with the
upper bound it caches this as a lower bound, and returns $\emptyset$.
Otherwise it caches the result and returns it.

\section{Experiments}

The goal of this section is to show a) the individual importance of the components of our algorithm, and b) that we can compute provably optimal trees with respect to nonlinear metrics for a large number of datasets. Furthermore, we show the difference in tree quality when comparing trees obtained using a single-objective approach and our bi-objective method, demonstrating the trade-off between quality and runtime offered by our algorithm. 

\emph{Benchmarks and experimental setting}. We considered 75 binary classification datasets used in previous works \cite{verwer2019learning,dl85,murtree,narodytska2018learning,hu2019optimal}. The experiments were run one at a time on an Intel i7-3612QM@2.10 GHz with 8 GB RAM. 

\emph{Public release.} The code and benchmarks are available at \url{bitbucket.org/EmirD/murtree-bi-objective}.

\subsection{Individual Algorithmic Components}

We experiment with variations of our approach to understand the impact of
each technique. The default configuration of our algorithm uses all
techniques as presented in Figure \ref{fig:code}. We compare the default
configuration with three variations, each excluding a single technique:
(A) upper bounding (B) infeasibility-based lower bound,  (D)
similarity-based lower bounding. 
We use trees of depth four and observe the runtime to compute optimality. In
the interest of space, we provide results for selected datasets (Table \ref{table:small}) and briefly discuss the rest.
All three techniques consistently leads to runtime improvements. The exception is similarity-based lower bounding on two datasets,
which is a caveat of our current implementation, i.e., it recomputes the same lower bound unnecessarily in certain scenarios. We note that the dataset \emph{magic04} has an exceptionally large Pareto front compared, resulting in large overheads for techniques that manipulate the Pareto front. The significance of individual techniques heavily depends on the dataset, but overall the techniques synergise, since lower and upper bounds closely interact. We note that, while the techniques provide improvements, their advantage is not as pronounced as in the single-objective case. In general, bi-objective problems are more difficult to solve than their single-objective counterpart and we may expect weaker pruning mechanisms. Conditions for pruning are harder to achieve, e.g., \emph{every} element of the upper bound must be dominated by the lower bound for upper bound (A) pruning. The results on the remaining datasets is similar.
 
\input{table_small}

\subsection{Comparison to Single-Objective Optimisation}

The aim is to experimentally evaluate the generalisation properties of optimal F1-score trees, i.e., do optimal F1-score trees on the training set generalise to the test set? As there are no other approaches that compute the Pareto front or optimise nonlinear metrics, we compare with the baseline algorithm MurTree \cite{murtree}, a single-objective approach that optimises misclassifications. We consider F1-score as the nonlinear metric, but other metrics satisfying monotonicity (Eq. \ref{eq:monotonicity}) can be used without effecting runtime as computing the Pareto front is the bottleneck. 

We perform hyper-parameter tuning considering parameters $depth \in \{1, 2, 3, 4\}$ and $size \in \{1, 2, ..., 2^{depth}-1\}$. Five-fold cross-validation is used to evaluate each combination of parameters and the parameters that maximises accuracy or F1-score on test set across the folds is selected. The timeout is set to one hour for each benchmark. We considered 75 datasets and results are shown in Table \ref{table:big}. For brevity, we removed datasets that were computed within a second, and those that neither the single- or bi-objective technique managed to fully tune within the time limit. There are three main conclusions:

\begin{itemize}
\item There is value in computing optimal trees with respect to a nonlinear metric. Indeed, trees computed by optimising directly the F1-score have equal or higher scores on the train and test sets that run within the time limit. The extent of the benefit depends on the nature of the dataset. Naturally, imbalanced datasets are more likely to benefit from nonlinear metrics, e.g., fico, seismic\_bumps, bank\_conv, IndiansDiabetes, appendicitis-un, german-credit. A prime example of an imbalanced dataset is \emph{seismic}\_\emph{bumps}, i.e., 170 positives instances compared to 2414 negative instances, where the difference in F1 score among the tree trained with accuracy and the tree train with F1 is significant (e.g., 0.02 vs 0.31).

\item The difference between tuning on accuracy and f1-score is exemplified on the depth-three trees compared to depth-four trees. This is inline with expectations, as shallower trees optimised with accuracy are more likely to be biased towards the majority class, resulting in lower F1-scores (depth-zero trees are the extreme example).

\item The runtime is higher to compute optimal trees with respect to a nonlinear metric than using accuracy. This is expected given nonlinear problems are notably harder to solve than problems with linear objectives.
\end{itemize}

Note that while the runtime of hyper-parameter tuning and cross-validating may be high, classifying new samples can be done quickly. Depending on the application, the high training times may not be a considerable drawback.  This represents a trade-off that must be evaluated for a given application. Our results show that for cases where a nonlinear metric is desirable, provided there is enough time, it is beneficial to optimise the metric directly using our approach.

\input{table_big}

\section{Related Work}

Decision trees are traditionally constructed using heuristic algorithms, e.g., CART algorithm \cite{breiman1984classification}. While scalable, the resulting tree may not be the most accurate tree. \emph{Optimal} decision tree algorithms, which are the focus of this work, combat this issue by exhaustively exploring the space of decision trees, at the expense of runtime.

One research line for optimal decision tree uses generic optimisation approaches. The main idea is that a decision tree, together with the data, can be stated in terms of binary variables and constraints as a mathematical program, such that each feasible solution to the mathematical formulation can be translated to a decision tree. The objective function captures the desired metrics, e.g., accuracy. 

Recently, \citeay{bertsimas2017optimal} and \citeay{verwer2017learning} proposed novel mixed-integer programming formulations. The methods encode the optimal decision tree by fixing the tree depth in advance, creating variables to represent the predicates for each node, and adding constraints to enforce the decision tree structure. These approaches were later improved by \emph{BinOPT}~\cite{verwer2019learning}, a \emph{binary linear programming} formulation, that took advantage of implicitly binarising data to reduce the number of variables and constraints required to encode the problem. \citeay{aghaei2019learning} encoded fairness metrics within a mixed-integer programming formulation for optimal decision trees in socially sensitive contexts, which aims to deliver \emph{fair} and \emph{accurate} decision trees. Note that while integer programming has been used to optimise accuracy, it is not straight-forward to modify integer programming methods to include nonlinear metrics since the objective function is nonlinear.

\citeauthor{narodytska2018learning}~(\citeyear{narodytska2018learning}) used an encoding of decision trees into propositional logic (SAT) to construct the smallest tree in terms of the total number of nodes that \emph{perfectly} describes the given dataset, i.e., leads to zero misclassifications on the training data. An initial perfect decision tree is constructed using a heuristic method, after which a series of SAT-solver calls are made, each time posing the problem of computing a perfect tree with one less node. The SAT approach of \cite{avellanedaefficient} simplifies the encoding by fixing the depth of the tree and employing an incremental approach to gradually added instances to the formulation.

\citeauthor{verhaeghe2019learning}~(\citeyear{verhaeghe2019learning}) approached the optimal classification tree problem by minimising the misclassifications using constraint programming. The approach captured the decomposable structure of decision trees within an AND-OR search framework, i.e., once a feature for a parent node has been selected, the child nodes can be optimised independently. Upper bounding on the number of misclassifications was used to prune parts of the search space and their algorithm incorporated an itemset mining technique to speed-up the computation of instances per node and used a caching technique similar to \emph{DL8} (see below).

Another stream of research develops tailored algorithms for decision trees. \citeay{dl8} introduced a framework named \emph{DL8} for optimal decision trees that could support a wide range of constraints. Their algorithm took advantage that the left and right subtree of a given node can be optimised independently, introduced a caching technique to save subtrees computed during the algorithm in order to reuse them at a later stage, and combined these with ideas from the pattern mining literature to compute optimal decision trees. DL8 \cite{dl8} laid important algorithmic foundations for optimal decision trees.

\citeay{hu2019optimal} presented an algorithm that computes the optimal decision tree by considering a balance between misclassifications and number of nodes. They apply exhaustive search, caching, and lower bounding of the misclassifications based on the cost of adding a new node to the decision tree. Compared to other recent optimal decision tree algorithms, the method relies on the number of nodes playing an important role in the metric of optimality and a limited number of binary features, e.g., the authors experimented with datasets with up to twelve binary features. \cite{icmlTree} improved the algorithm and added support for additional metrics, however the F1-score was not directly optimised but rather linearised.

\citeay{dl85} developed \emph{DL8.5} by combining and refining the ideas from \emph{DL8} and the constraint programming approach. The approach computes the most accurate decision tree with respect to a constraint on the depth of the tree and has been made available as a Python library \cite{aglin2020pydl8}. The main addition was an upper bound pruning technique, which limited the upper misclassification value of a child node once the optimal subtree was computed for its sibling, and a lowering bound technique, where the algorithm stored information not only about computed optimal subtrees but also pruned subtrees to provide a lower bound on the misclassifications of a subtree. \cite{murtree} advanced the DL8.5 algorithm by adding support to limit the number of nodes in the tree, an efficient procedure to compute tree of depth two, and a novel similarity-based lower bounding approach. Recall that these approaches have been presented in the preliminary section. 

The main lesson learned in previous works is that exploiting properties specific to decision trees leads to substantial gains in performance, regardless of whether it was applied in a dedicated or generic optimisation approach.

For completeness, we note that there are other works that follow a different but related research line, which use neural networks to learn decision trees/forests \cite{tanno2019adaptive,kontschieder2015deep}, consider more general tree structures \cite{yang2019weighted}, or end-to-end learning using decision trees \cite{hehn2019end,elmachtoub2020decision}. Given that our aim diverges from these works, we do not further discuss them.

\section{Conclusion}

Nonlinear metrics are generally agreed as better methods for evaluating
the performance machine learning models on imbalanaced datasets. 
We provide the first approach to generate decision trees which are optimal
under a monotonic nonlinear metric.  
We show that existing linear metric approaches do not yield optimal decision
trees under nonlinear metrics.
The approach can readily be extended to
generating sparse trees that tradeoff (linear) size versus the non-linear
metric by generating Pareto frontiers for each size and choosing the best,
though exploration of more efficient approaches seems worthwhile, in particular, stronger bounding techniques.

\bibliography{references}

\end{document}

%% file: table_small.tex
\begin{table*}
\centering
\tiny
\def\d{@{\hspace*{1.7mm}}}
\resizebox{\textwidth}{!}{
\begin{tabular}{@{}  l | r | r | r | r | r || r | r | r || r @{} }
	name &  $|\mathcal{D}|$ & $|\mathcal{F}|$ &  $|\mathcal{D^+}|$ &  $|\mathcal{D^-}|$ &  $|PF|$ &  $-$(A)  &  $-$(B) &  $-$(D) & default \\ \hline
	anneal & 812 & 93 & 625 & 187 & 63         & 20          & 18  & 35           & \textbf{15} \\ 
	au-credit & 653 & 125 & 357 & 296 & 54     & 85          & 85  & 90           & \textbf{78} \\
	breast-wis & 683 & 120 & 444 & 239 & 8     & 28          & 28  & 38           & \textbf{25} \\ 
	diabetes & 768 & 112 & 500 & 268 & 122     & 118         & 109 & 93           & \textbf{73} \\ 
	de-credit & 1000 & 112 & 700 & 300 & 185   & 184         & 180 & \textbf{140} & \textbf{140} \\
	heart-c & 296 & 95 & 160 & 136 & 27        & 26          & 27  & 29           & \textbf{24} \\ 
	kr-vs-kp & 3196 & 73 & 1669 & 1527 & 72    & 17          & 16  & 18           & \textbf{13} \\ 
	yeast & 1484 & 89 & 463 & 1021 & 296       & 232         & 250 & \textbf{130} & 156 \\ 
	bank & 4521 & 26 & 521 & 4000 & 364        & 17          & 17  & 10           & \textbf{4} \\
	HTRU\_2 & 17898 & 57 & 1639 & 16259 & 324 & 54          & 57  & \textbf{33}  & 37 \\ 
	Iono. & 351 & 143 & 225 & 126 & 7          & 79          & 91  & 110          & \textbf{61} \\ 
	magic04 & 19020 & 79 & 6688 & 12332 & 1408 & \textbf{98} & 100 & 104          & 905 \\
	spambase & 4601 & 132 & 1813 & 2788 & 321  & 516         & 492 & \textbf{360} & 411 \\ 
\end{tabular}
}
\caption{Runtime (sec) of variations by disabling a single technique (similarity-based lower bounding, upper bounding, and infeasibility lower bounds) on selected datasets. The size of the Pareto front is labelled as $|PF|$. Bold indicates the best result.\label{table:small}}%
\end{table*}

%% file: table_big.tex
\begin{table*}
\centering
\def\d{@{\hspace*{1.7mm}}}
\resizebox{2\columnwidth}{!}{
\begin{tabular}{ l | r | r | r | r | r | r | r | r | r | r | r | r | r | r | r | c}
	 & & \  & \  & & \multicolumn{6}{c|}{DEPTH = 3} & \multicolumn{6}{c}{DEPTH = 4}  \\
	   &    &    &    &  & \multicolumn{3}{|c|}{Accuracy-Tuned}   & \multicolumn{3}{|c|}{F1-Score-Tuned}  &  \multicolumn{3}{|c|}{Accuracy-Tuned} &   \multicolumn{3}{|c}{F1-Score-Tuned}  \\
	name &   $|\mathcal{D}|$ &   $|\mathcal{D}^+|$ &   $|\mathcal{D}^-|$ &   $|\mathcal{F}|$ &   train &   test &   time &   train &   test &   time &   train &   test &   time &   train &   test &   time \\  \hline

 compas-binary  &  6907  &  3196  &  3711  &  12  &  0.63  &  0.64  &  $<1s$  &  \textbf{ 0.67}  &  \textbf{ 0.68}  &  23  &  0.67  &  0.67  &  3  &  ---  &  ---  &  $>1h$  \\ 
 fico-binary  &  10459  &  5000  &  5459  &  17  &  0.69  &  0.69  &  1  &  \textbf{ 0.71}  &  \textbf{ 0.72}  &  69  &  0.72  &  0.72  &  11  &  ---  &  ---  &  $>1h$  \\ 
 banknote  &  1372  &  610  &  762  &  16  &  \textbf{ 0.91}  &  \textbf{ 0.92}  &  $<1s$  &  \textbf{ 0.91}  &  \textbf{ 0.92}  &  1  &  \textbf{ 0.94}  &  \textbf{ 0.94}  &  9  &  0.93  &  \textbf{ 0.94}  &  10  \\ 
 bank  &  4521  &  521  &  4000  &  26  &  0.50  &  0.52  &  $<1s$ &  \textbf{ 0.55}  &  \textbf{ 0.56}  &  10  &  0.50  &  0.52  &  1  &  \textbf{ 0.55}  &  \textbf{ 0.56}  &  2262  \\ 
 biodeg  &  1055  &  356  &  699  &  81  &  \textbf{ 0.76}  &  \textbf{ 0.77}  &  $<1s$  &  \textbf{ 0.76}  &  \textbf{ 0.78}  &  23  &  0.88  &  0.87  &  40  &  ---  &  ---  &  $>1h$  \\ 
 HTRU\_2  &  17898  &  1639  &  16259  &  57  &  \textbf{ 0.87}  &  \textbf{ 0.87}  &  3  &  \textbf{ 0.87}  &  \textbf{ 0.87}  &  69  &  0.98  &  0.98  &  143  &  ---  &  ---  &  $>1h$  \\ 
 IndiansDiabetes  &  768  &  268  &  500  &  11  &  0.65  &  0.70  &  $<1s$  &  \textbf{ 0.68}  &  \textbf{ 0.73}  &  1  &  0.65  &  0.70  &  1  &  \textbf{ 0.68}  &  \textbf{ 0.73}  &  55  \\ 
 Ionosphere  &  351  &  225  &  126  &  143  &  \textbf{ 0.93}  &  \textbf{ 0.92}  &  1  &  \textbf{ 0.93}  &  \textbf{ 0.92}  &  17  &  \textbf{ 0.92}  &  \textbf{ 0.93}  &  60  &  \textbf{ 0.92}  &  \textbf{ 0.93}  &  1247  \\ 
 magic04  &  19020  &  6688  &  12332  &  79  &  \textbf{ 0.68}  &  \textbf{ 0.69}  &  5  &  \textbf{ 0.68}  &  \textbf{ 0.69}  &  198  &  0.82  &  0.82  &  287  &  ---  &  ---  &  $>1h$  \\ 
 messidor  &  1151  &  611  &  540  &  24  &  0.67  &  0.70  &  $<1s$  &  \textbf{ 0.70}  &  \textbf{ 0.71}  &  7  &  0.69  &  0.71  &  2  &  \textbf{ 0.72}  &  \textbf{ 0.72}  &  1714  \\ 
 monk2  &  169  &  64  &  105  &  15  &  0.56  &  0.56  &  $<1s$  &  \textbf{ 0.67}  &  \textbf{ 0.68}  &  1  &  0.56  &  0.56  &  1  &  \textbf{ 0.78}  &  \textbf{ 0.69}  &  20  \\ 
 monk3  &  122  &  60  &  62  &  15  &  \textbf{ 0.93}  &  \textbf{ 0.94}  &  $<1s$ &  \textbf{ 0.93}  &  \textbf{ 0.94}  &  1  &  \textbf{ 0.93}  &  \textbf{ 0.96}  &  1  &  \textbf{ 0.93}  &  \textbf{ 0.96}  &  3  \\ 
 seismic\_bumps  &  2584  &  170  &  2414  &  10  &  0.02  &  0.04  &  $<1s$  &  \textbf{ 0.32}  &  \textbf{ 0.36}  &  2  &  0.02  &  0.04  &  1  &  \textbf{ 0.31}  &  \textbf{ 0.36}  &  63  \\ 
 spambase  &  4601  &  1813  &  2788  &  132  &  \textbf{ 0.86}  &  \textbf{ 0.87}  &  4  &  \textbf{ 0.86}  &  \textbf{ 0.87}  &  110  &  0.91  &  0.91  &  314  &  ---  &  ---  &  $>1h$  \\ 
 tic-tac-toe  &  958  &  626  &  332  &  18  &  0.62  &  0.64  &  $<1s$  &  \textbf{ 0.68}  &  \textbf{ 0.67}  &  1  &  \textbf{ 0.77}  &  0.75  &  1  &  \textbf{ 0.77}  &  \textbf{ 0.77}  &  75  \\ 
 appendicitis-un  &  106  &  21  &  85  &  530  &  0.46  &  0.31  &  4  &  \textbf{ 0.53}  &  \textbf{ 0.42}  &  202  &  0.86  &  0.84  &  190  &  ---  &  ---  &  $>1h$  \\ 
 cancer-un  &  683  &  239  &  444  &  89  &  \textbf{ 0.95}  &  \textbf{ 0.94}  &  $<1s$  &  \textbf{ 0.95}  &  \textbf{ 0.94}  &  5  &  \textbf{ 0.95}  &  \textbf{ 0.94}  &  11  &  \textbf{ 0.95}  &  \textbf{ 0.94}  &  313  \\ 
 car-un  &  1728  &  518  &  1210  &  21  &  \textbf{ 0.83}  &  0.83  &  $<1s$  &  \textbf{ 0.83}  &  \textbf{ 0.84}  &  1  &  0.85  &  0.86  &  2  &  \textbf{ 0.86}  &  \textbf{ 0.87}  &  58  \\ 
 cleve-un  &  303  &  138  &  165  &  395  &  0.82  &  0.82  &  3  &  \textbf{ 0.83}  &  \textbf{ 0.83}  &  236  &  0.86  &  0.83  &  221  &  ---  &  ---  &  $>1h$  \\ 
 colic-un  &  368  &  232  &  136  &  415  &  \textbf{ 0.80}  &  \textbf{ 0.82}  &  7  &  \textbf{ 0.80}  &  \textbf{ 0.82}  &  304  &  0.86  &  0.85  &  670  &  ---  &  ---  &  $>1h$  \\ 
 corral-un  &  160  &  70  &  90  &  6  &  \textbf{ 0.94}  &  \textbf{ 0.92}  &  $<1s$  &  \textbf{ 0.94}  &  \textbf{ 0.92}  &  1  &  \textbf{ 1.00}  &  \textbf{ 1.00}  &  1  &  \textbf{ 1.00}  &  \textbf{ 1.00}  &  0.10  \\ 
 haberman-un  &  306  &  225  &  81  &  92  &  0.29  &  0.25  &  $<1s$ &  \textbf{ 0.49}  &  \textbf{ 0.54}  &  8  &  0.29  &  0.25  &  7  &  \textbf{ 0.49}  &  \textbf{ 0.55}  &  1030  \\ 
 heart-statlog-un  &  270  &  120  &  150  &  381  &  0.81  &  0.83  &  3  &  \textbf{ 0.84}  &  \textbf{ 0.84}  &  202  &  0.87  &  0.86  &  183  &  ---  &  ---  &  $>1h$  \\ 
 hepatitis-un  &  155  &  32  &  123  &  361  &  0.45  &  0.50  &  4  &  \textbf{ 0.55}  &  \textbf{ 0.63}  &  120  &  0.85  &  0.84  &  236  &  ---  &  ---  &  $>1h$  \\ 
 house-votes-84-un  &  435  &  267  &  168  &  16  &  \textbf{ 0.94}  &  \textbf{ 0.95}  &  $<1s$  &  \textbf{ 0.94}  &  \textbf{ 0.95}  &  1  &  \textbf{ 0.94}  &  \textbf{ 0.95}  &  1  &  \textbf{ 0.94}  &  \textbf{ 0.95}  &  6  \\ 
 hungarian-un  &  294  &  106  &  188  &  330  &  0.77  &  0.80  &  4  &  \textbf{ 0.78}  &  \textbf{ 0.81}  &  134  &  0.82  &  0.81  &  210  &  ---  &  ---  &  $>1h$  \\ 
 mouse-un  &  70  &  29  &  41  &  45  &  \textbf{ 0.96}  &  \textbf{ 0.90}  &  $<1s$  &  \textbf{ 0.96}  &  \textbf{ 0.90}  &  1  &  \textbf{ 0.96}  &  \textbf{ 0.90}  &  1  &  \textbf{ 0.96}  &  \textbf{ 0.90}  &  4  \\ 
 promoters-un  &  106  &  53  &  53  &  334  &  \textbf{ 0.97}  &  0.86  &  6  &  0.87  &  \textbf{ 0.87}  &  125  &  0.96  &  0.83  &  467  &  ---  &  ---  &  $>1h$  \\ 
 spect-un  &  267  &  212  &  55  &  22  &  0.42  &  0.50  &  $<1s$ &  \textbf{ 0.59}  &  \textbf{ 0.66}  &  1  &  0.42  &  0.48  &  1  &  \textbf{ 0.58}  &  \textbf{ 0.66}  &  37  \\ 
 anneal  &  812  &  625  &  187  &  93  &  0.64  &  0.66  &  $<1s$  &  \textbf{ 0.66}  &  \textbf{ 0.70}  &  11  &  \textbf{ 0.73}  &  0.72  &  12  &  \textbf{ 0.73}  &  \textbf{ 0.75}  &  1397  \\ 
 audiology  &  216  &  57  &  159  &  148  &  0.91  &  \textbf{ 0.93}  &  $<1s$  &  \textbf{ 0.92}  &  \textbf{ 0.93}  &  6  &  0.91  &  \textbf{ 0.93}  &  12  &  \textbf{ 0.95}  &  \textbf{ 0.93}  &  429  \\ 
 australian-credit  &  653  &  357  &  296  &  125  &  0.86  &  \textbf{ 0.88}  &  $<1s$  &  \textbf{ 0.87}  &  \textbf{ 0.88}  &  29  &  0.86  &  \textbf{ 0.88}  &  50  &  \textbf{ 0.87}  &  \textbf{ 0.88}  &  3418  \\ 
 breast-wisconsin  &  683  &  444  &  239  &  120  &  \textbf{ 0.96}  &  \textbf{ 0.96}  &  $<1s$  &  \textbf{ 0.96}  &  \textbf{ 0.96}  &  9  &  \textbf{ 0.96}  &  \textbf{ 0.97}  &  22  &  \textbf{ 0.96}  &  \textbf{ 0.97}  &  584  \\ 
 diabetes  &  768  &  500  &  268  &  112  &  0.68  &  0.71  &  $<1s$  &  \textbf{ 0.70}  &  \textbf{ 0.72}  &  44  &  0.77  &  0.77  &  53  &  ---  &  ---  &  $>1h$  \\ 
 german-credit  &  1000  &  700  &  300  &  112  &  0.54  &  0.57  &  $<1s$  &  \textbf{ 0.60}  &  \textbf{ 0.64}  &  71  &  0.75  &  0.73  &  60  &  ---  &  ---  &  $>1h$  \\ 
 heart-cleveland  &  296  &  160  &  136  &  95  &  \textbf{ 0.82}  &  \textbf{ 0.82}  &  $<1s$  &  \textbf{ 0.82}  &  \textbf{ 0.82}  &  12  &  \textbf{ 0.82}  &  \textbf{ 0.82}  &  16  &  \textbf{ 0.82}  &  \textbf{ 0.82}  &  911  \\ 
 hepatitis  &  137  &  111  &  26  &  68  &  0.59  &  \textbf{0.61}  &  $<1s$  &  0.55  &  \textbf{ 0.67}  &  2  &  \textbf{ 0.59}  &  0.61  &  3  &  0.55  &  \textbf{ 0.67}  &  100  \\ 
 hypothyroid  &  3247  &  2970  &  277  &  88  &  \textbf{ 0.89}  &  \textbf{ 0.90}  &  2  &  \textbf{ 0.89}  &  \textbf{ 0.90}  &  12  &  \textbf{ 0.90}  &  \textbf{ 0.91}  &  48  &  \textbf{ 0.90}  &  \textbf{ 0.91}  &  1070  \\ 
 kr-vs-kp  &  3196  &  1669  &  1527  &  73  &  \textbf{ 0.93}  &  \textbf{ 0.93}  &  2  &  \textbf{ 0.93}  &  \textbf{ 0.93}  &  8  &  \textbf{ 0.95}  &  \textbf{ 0.95}  &  38  &  \textbf{ 0.95}  &  \textbf{ 0.95}  &  849  \\ 
 lymph  &  148  &  81  &  67  &  68  &  \textbf{ 0.90}  &  \textbf{ 0.86}  &  $<1s$  &  \textbf{ 0.90}  &  \textbf{ 0.86}  &  2  &  \textbf{ 0.95}  &  0.86  &  3  &  \textbf{ 0.95}  &  \textbf{ 0.87}  &  115  \\ 
 mushroom  &  8124  &  4208  &  3916  &  119  &  \textbf{ 1.00}  &  \textbf{ 1.00}  &  4  &  \textbf{ 1.00}  &  \textbf{ 1.00}  &  9  &  \textbf{ 1.00}  &  \textbf{ 1.00}  &  121  &  \textbf{ 1.00}  &  \textbf{ 1.00}  &  416  \\ 
 pendigits  &  7494  &  780  &  6714  &  216  &  0.96  &  \textbf{ 0.97}  &  47  &  \textbf{ 0.97}  &  \textbf{ 0.97}  &  134  &  \textbf{1.0}  &  \textbf{1.0}  &  2718  &  ---  &  ---  &  $>1h$  \\ 
 primary-tumor  &  336  &  82  &  254  &  31  &  0.65  &  0.66  &  $<1s$  &  \textbf{ 0.64}  &  \textbf{ 0.68}  &  2  &  \textbf{ 0.65}  &  0.66  &  1  &  0.64  &  \textbf{ 0.68}  &  100  \\ 
 segment  &  2310  &  330  &  1980  &  235  &  \textbf{ 0.99}  &  \textbf{ 1.00}  &  8  &  \textbf{ 0.99}  &  \textbf{ 1.00}  &  30  &  \textbf{ 0.99}  &  \textbf{ 1.00}  &  154  &  \textbf{ 0.99}  &  \textbf{ 1.00}  &  1832  \\ 
 soybean  &  630  &  92  &  538  &  50  &  0.81  &  \textbf{ 0.88}  &  $<1s$  &  \textbf{ 0.82}  &  \textbf{ 0.88}  &  2  &  0.82  &  0.91  &  4  &  \textbf{ 0.86}  &  \textbf{ 0.92}  &  87  \\ 
 tic-tac-toe  &  958  &  626  &  332  &  27  &  0.61  &  0.63  &  $<1s$  &  \textbf{ 0.68}  &  \textbf{ 0.67}  &  3  &  \textbf{ 0.78}  &  \textbf{ 0.78}  &  4  &  \textbf{ 0.78}  &  \textbf{ 0.78}  &  242  \\ 
 vehicle  &  846  &  218  &  628  &  252  &  \textbf{ 0.93}  &  \textbf{ 0.94}  &  8  &  \textbf{ 0.93}  &  \textbf{ 0.94}  &  70  &  0.97  &  0.97  &  600  &  ---  &  ---  &  $>1h$  \\ 
 vote  &  435  &  267  &  168  &  48  &  \textbf{ 0.94}  &  \textbf{ 0.96}  &  $<1s$  &  \textbf{ 0.94}  &  \textbf{ 0.96}  &  1  &  \textbf{ 0.94}  &  \textbf{ 0.96}  &  5  &  \textbf{ 0.94}  &  \textbf{ 0.96}  &  59  \\ 
 yeast  &  1484  &  463  &  1021  &  89  &  0.45  &  0.49  &  1  &  \textbf{ 0.60}  &  \textbf{ 0.63}   &  66  &  0.74  &  0.73  &  60  &  ---  &  ---  &  $>1h$  \\ 
\end{tabular}
}
\caption{The F1-score is displayed for depth three and four trees with optimal accuracy and optimal F1-score and the time to hyper-parameter tune with five-fold cross validation. Trees optimised with F1-score lead to better out-of-sample scores albeit at higher runtime. Time limit is set to one hour. Timeouts denoted as $-$. For each dataset, we show the number of instances ($|\mathcal{D}|$) and number of binary features ($\mathcal{F})$. Bold indicates the best result.\label{table:big} }%
\end{table*}